\documentclass[a4paper,11pt,twoside,onecolumn]{llncs}

\usepackage[utf8]{inputenc}

\begin{document}

\title{Towards Using Machine Translation Techniques to Induce Multilingual Lexica of Discourse Markers}
%\subtitle{-- Extended Abstract --}

\author{António Lopes\inst{1,2} \and David Martins de Matos\inst{1,2} \and Vera Cabarrão\inst{1,4} \and Ricardo Ribeiro\inst{1,3} \and Helena Moniz\inst{1} \and Isabel Trancoso\inst{1,2} \and Ana Isabel Mata\inst{4}\\
\email{antonio.lopes \and david.matos \and vera.cabarrao \and ricardo.ribeiro \and helena.moniz (@l2f.inesc-id.pt)}
}
\institute{L2F/INESC-ID, Rua Alves Redol 9, 1000-029 Lisboa, Portugal
\and IST/Universidade de Lisboa, Av. Rovisco Pais, 1, 1049-001 Lisboa, Portugal
\and ISCTE-IUL, Av. das Forças Armadas, 1649-026 Lisbon, Portugal 
\and FL/Universidade de Lisboa, Alameda da Universidade, 1600-214 Lisbon, Portugal\\
}

%\email{,,isabel.trancoso,anaisabelmata}

\maketitle

\section{Introduction}

Discourse markers are universal linguistic events subject to language variation. Although an extensive literature has already reported language specific traits of these events (e.g.~\cite{Fraser1990,Fraser1999,Fischer2000,Beeching2014,Ghezzi2014}), little has been said on their cross-language behavior and, subsequently, on building an inventory of multilingual lexica of discourse markers. Thus, this work describes new methods and approaches for the description, classification, and annotation of discourse markers in the specific domain of the Europarl corpus. The study of discourse markers in the context of translation is crucial due to the idiomatic nature of these structures (e.g.~\cite{Aijmer2007,Beeching2013}). Multilingual lexica together with the functional analysis of such structures are useful tools for the hard task of translating discourse markers into possible equivalents from one language to another.

%\subsection{Objective}

Using Daniel Marcu's validated discourse markers for English~\cite{Marcu2000}, extracted from the Brown Corpus~\cite{Francis1979}, our purpose is to build multilingual lexica of discourse markers for other languages, based on machine translation techniques.
The major assumption in this study is that the usage of a discourse marker is independent of the language, i.e., the rhetorical function of a discourse marker in a sentence in one language is equivalent to the rhetorical function of the same discourse marker in another language.

\section{Methodology}

We used the European Parliament corpus\footnote{Available at \url{http://www.statmt.org/europarl/} (visited March 2015).}, version 7, in this experiment. The procedure is applied to all the pairs of languages available. The corpus consists of proceedings of the European Parliament. It includes versions in 21 European languages: Romanic (French, Italian, Spanish, Portuguese, Romanian), Germanic (English, Dutch, German, Danish, Swedish), Slavik (Bulgarian, Czech, Polish, Slovak, Slovene), Finni-Ugric (Finnish, Hungarian, Estonian), Baltic (Latvian, Lithuanian), and Greek.

Machine translation systems can be classified according to the atomic units to be translated: for example, while for word-based methods the atomic unit is the word, for phrase-based methods the atomic unit is the phrase. Thus, the most important knowledge sources of phrase-based methods are tables of possible phrase translations between language pairs. Phrase-based methods, due to their nature, have, at least, one interesting advantage for this specific work: they can handle non-compositional phrases.

Before computing the phrase table for each language pair in the corpus, alignment and normalization steps are necessary. These steps ensure that each phrase in one language has a counterpart in the other language. The alignment algorithm is an implementation of the Church and Gale algorithm for bilingual corpora~\cite{Gale1993}. However, in this step, the algorithm was adapted to take into account specific aspects of the Europarl corpus. This was necessary because the least alignment unit is the paragraph, not the phrase. In this case, when the number of lines per paragraph is different in the two languages, all the paragraphs are collapsed into a single line. After the previous step, a normalization step is carried out. First, all SGML (Standard Generalized Markup Language) tags are removed, since they are no longer needed in the aligned corpus; second, sentences are tokenized (using a tool provided by the Europarl package that constitutes a weak language dependency, i.e., it only separates word tokens); and, third, all text is converted to lowercase.

\subsection{Building the phrase table}

In order to create the phrase table between foreign-English language pairs, for all the experiments, we used the Moses decoder\footnote{\url{http://www.statmt.org/moses/index.php?n=Main.HomePage}}, particularly the train model, with the default parameters. This function uses the GIZA++ tool~\cite{Och2003}, an implementation of the IBM models, and is used to establish word alignments. 

\subsection{Pruning the phrase table}

As the phrase table contains all pairs found in the parallel corpus, there is much noise and, consequently, not all pairs are likely to be either selected as candidate or as a good marker translation. This occurs because the models take into account all possible observations. We used a tool provided by the Moses Decoder that re-implements the algorithm proposed in~\cite{Johnson2007}, which prunes out unlikely pairs. This tool is dependent of the SALM tool ~\cite{Zhang2006}.

\subsection{Selection of discourse markers candidates}

After the creation of the phrase table, we developed a method for extracting from the table only the desired translations. This was accomplished by searching the table for each of the target discourse markers, where the marker appears followed or preceded by any kind of punctuation marks.

\subsection{Filtering undesirable translations}

Since we want the best translations of the markers, we developed a way to filter and remove all the undesirable candidates generated in the previous step. The filtering process uses the information from the phrase table: the translation to the target language, the marker in English, the inverse phrase translation probability $\varphi(f|e)$, the inverse lexical weighting $lex(f|e)$, the direct phrase translation probability $\varphi(e|f)$, the direct lexical weighting $lex(e|f)$, and the word-to-word alignment. Beyond the phrase table information, several heuristics were used to help filtering out bad alignments.

\section{Preliminary Results}

The following table presents some translation examples for markers above all and since to different languages.

\begin{table}
\centering
\begin{tabular}{|p{.15\columnwidth}|p{.2\columnwidth}|p{.22\columnwidth}|p{.15\columnwidth}|p{.2\columnwidth}|}\hline
\textbf{English} & \textbf{Portuguese}      & \textbf{French} & \textbf{German} & \textbf{Italian} \\\hline
above all & sobretudo       & avant tout & vor allem & soprattutto \\
          & sobretudo de    & surtout    & vor allen & soprattutto di \\
          & acima de tudo   &            &           & \\
          & acima de todas  &            &           & \\\hline
since     & pois            & puisque    & da        & dal momento \\
          & desde então     & depuis            & seit      & dal \\
          & desde de        & dans la mesure où & da seit   & poiché \\
          & desde que       &                   & die seit  & \\
          &                 &                   & seit sich & \\\hline
\end{tabular}
\end{table}

The original list of 427 markers in English generated for the languages selected for the above example, 846 for Portuguese (27 were pruned from the phrase table), 861 for French (40 removed after pruning), 906 for German (43 do not exist in the phrase table after pruning), and 1293 for Italian (46 do not exist in the phrase table after pruning).

Another experiment aiming at a cross-domain analysis of discourse markers in two spontaneous speech corpora (university lectures and map-task dialogues) was conducted to identify and classify discourse markers in Portuguese. The discourse markers collected were coded as conversational markers, i.e., those that only occur in oral communications, and as both conversational and textual markers, meaning those that can occur both in speech and in written texts. In order to compare the discourse markers in these corpora with the Portuguese lexicon of discourse markers in the Europarl corpus, extracted within this work, we searched their equivalents in the latter for English. Results showed that, out of around 70 discourse markers, only 18 were available and had an English translation. As for the classification of the discourse markers found in the Europarl corpus, 7 were coded both as conversational and textual markers, and the remaining were classified only as conversational markers. A possible interpretation for these results is that the register used in the three corpora differs considerably. University lectures and map-task dialogues have a much more informal type of speech than Europarl. The following table presents some examples.

\begin{table}
\centering
\begin{tabular}{|l|l|}\hline
\textbf{Portuguese}      & \textbf{English} \\\hline
A seguir   & After that; after; afterwards; following; next; thereafter \\
Agora      & From now on; now \\
Bem        & All right; and; as well; fine; okay; well \\
Bom        & Well \\
E depois   & And then; but then; then \\
Enfim      & Anyway; at last; finally; in short; lastly \\
Entretanto & But; in the meantime; meanwhile \\
Muito bem  & All right; fine; okay \\
Ora        & But; however; now; or; well; yet \\
Pois       & Because; on the grounds; since; then; therefore \\\hline
\end{tabular}
\end{table}

\section{Conclusions}

This preliminary work describes new methods and approaches for the description, classification, and annotation of discourse markers based on the Europarl corpus. Building multilingual lexica is a much need task, especially in the fields of machine translation and (computational) linguistics. As preliminary results show, multilingual lexica allow for the establishment of an inventory of discourse markers, for multiple translations of each entry in the inventory, and also for the analysis of contexts of usage in several languages. Per se, these lexica are a very useful tool to correlate cross-language discourse markers, validated by the fact that professional conference interpreters established those relations. Ultimately, this work may be considered as a step forward towards the rhetorical analysis of discourse markers in a cross-language framework.

\bibliographystyle{plain}
\bibliography{umtt}

\end{document}